\documentclass[sansbold=false]{phai}
\usepackage{tabularx}
\usepackage{algorithm}
\usepackage{algpseudocode}
\usepackage{placeins}
\usepackage{needspace}

\title{ScienceBuddy:\\Recursive-in-Recursive Self-Improvement\\for Interactive Scientific Agents}
\subtitle{}
\authors{%
Shuhan Xue\textsuperscript{1,*}\quad
Jianyuan Zhong\textsuperscript{1,*}\quad
Ziyuan Nan\textsuperscript{1,*}\quad
Wenbin Li\textsuperscript{1}\quad
Zhaochen Yu\textsuperscript{1}\\[0.6mm]
Jinchao Ding\textsuperscript{1}\quad
Qiang Gao\textsuperscript{2,3,4,5}\quad
Pengyu Zhan\textsuperscript{6}\quad
Yuntong Zhang\textsuperscript{6}\quad
Tian Cheng\textsuperscript{6}\\[0.6mm]
Zhenfei Yin\textsuperscript{1,7,\textdagger}\quad
Yingcheng Wu\textsuperscript{1,8,\textdagger}\quad
Ling Yang\textsuperscript{1,9,\textdagger}}
\affiliations{}
\legend{%
{\sffamily\fontsize{11}{13}\selectfont\color{phaiink}%
\textbf{Website}: \href{http://science-buddy.io}{Science-Buddy-Product}\quad\textcolor{phaifaint}{|}\quad
\textbf{Code}: \href{https://github.com/Gen-Verse/ScienceBuddy-RSI}{Gen-Verse/ScienceBuddy}}}
\checkdata[Corresponding]{yin@phai-labs.com; wuyc@phai-labs.com; yang@phai-labs.com}
\runningtitle{ScienceBuddy: Recursive-in-Recursive Self-Improvement}

\newcommand{\method}{ScienceBuddy}
\newcolumntype{Y}{>{\raggedright\arraybackslash}X}
\hypersetup{pdftitle={ScienceBuddy: Recursive-in-Recursive Self-Improvement for Interactive Scientific Agents},pdfauthor={Ziyuan Nan, Jianyuan Zhong, Shuhan Xue, Pengyu Zhan, Yuntong Zhang, Tian Cheng, Zhenfei Yin, Yingcheng Wu, Ling Yang},pdfsubject={Recursive-in-Recursive Self-Improvement for Interactive Scientific Agents}}

\input{frontmatter-layout}

\makeatletter
\renewcommand*{\l@paragraph}[2]{%
  \begingroup
    \normalfont\fontsize{9}{10.5}\selectfont
    \color{phaiink}%
    \@dottedtocline{4}{8em}{4.1em}{#1}{#2}%
  \endgroup
}
\makeatother

\begin{document}
\begingroup
\setlength{\parskip}{0pt}
\maketitle
\begingroup
\renewcommand{\thefootnote}{\fnsymbol{footnote}}
\footnotetext[1]{Equal contribution.}
\footnotetext[2]{Corresponding authors.}
\endgroup
\begin{abstract}

We introduce and release \textbf{ScienceBuddy}, an interactive scientific research workspace that brings continually improving scientific agents into researchers' everyday workflows. ScienceBuddy supports researchers in carrying out scientific tasks while transforming their requests, feedback, and execution evidence into tasks and evaluation rubrics for continual learning. At its core is \textbf{recursive-in-recursive self-improvement}, a paradigm that couples harness evolution with model reinforcement learning: the \textbf{inner recursion} improves the harness with the model fixed, while the \textbf{outer recursion} trains the model under the improved harness. Harness evolution shapes training experience, and model learning creates new opportunities for harness adaptation. We present case studies of researcher interaction, harness refinement, and model learning, with the benchmark cases spanning four scientific task families. By releasing ScienceBuddy as a research product, we make this paradigm available to the scientific community and take a step toward \textbf{discovery intelligence}: scientific AI that advances through sustained collaboration with researchers and evolves alongside the research it supports.

\end{abstract}
\noindent\hspace*{6mm}\begin{minipage}{\dimexpr\linewidth-12mm\relax}
\vspace{1mm}
{\linespread{1}\fontsize{9}{11.5}\selectfont\color{phaimuted}
\emph{``Agents will inhabit streams of experience, rather than short snippets of interaction.''}\par
\smallskip
{\raggedleft --- Silver and Sutton, \emph{Welcome to the Era of Experience} (2025)~\citep[p.~2]{eraofexperience}\par}}
\end{minipage}

\par


\enlargethispage{8mm}
\par\vspace{0.5mm}
\noindent\begin{minipage}{\linewidth}
\centering
\includegraphics[width=0.92\linewidth]{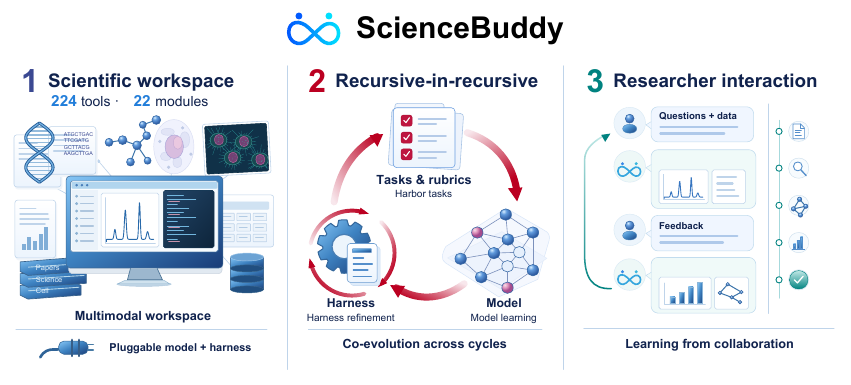}
\captionsetup{font=overviewcaption,skip=3pt,justification=justified,singlelinecheck=false}
\captionof{figure}{\textbf{ScienceBuddy: a scientific workspace that learns through collaboration.} \textbf{Left: Scientific workspace.} A multimodal workspace brings together documents, images, tables, and biological sequences with 224 tools across 22 functional modules, spanning genomics, molecular and cancer biology, pharmacology, bioimaging, literature retrieval, and database queries. Pluggable frontier models and agent harnesses support scientific analysis within this shared environment. \textbf{Middle: Recursive-in-recursive self-improvement.} Nested harness refinement and model learning are linked through scientific tasks and evaluation rubrics. \textbf{Right: Researcher interaction.} Researchers pose questions, inspect results, and refine requirements. These exchanges supply task objectives, evaluation criteria, and evidence for further improvement, connecting scientific collaboration to the next learning cycle.}
\label{fig:sciencebuddy-teaser}
\end{minipage}

\par\endgroup

\clearpage
\begin{figure}[p]
\centering
\makebox[\linewidth][c]{%
\begin{minipage}{160mm}
\centering
\setlength{\parskip}{0pt}
\captionsetup{font=overviewcaption}
\includegraphics[width=150mm,height=0.77\textheight,keepaspectratio]{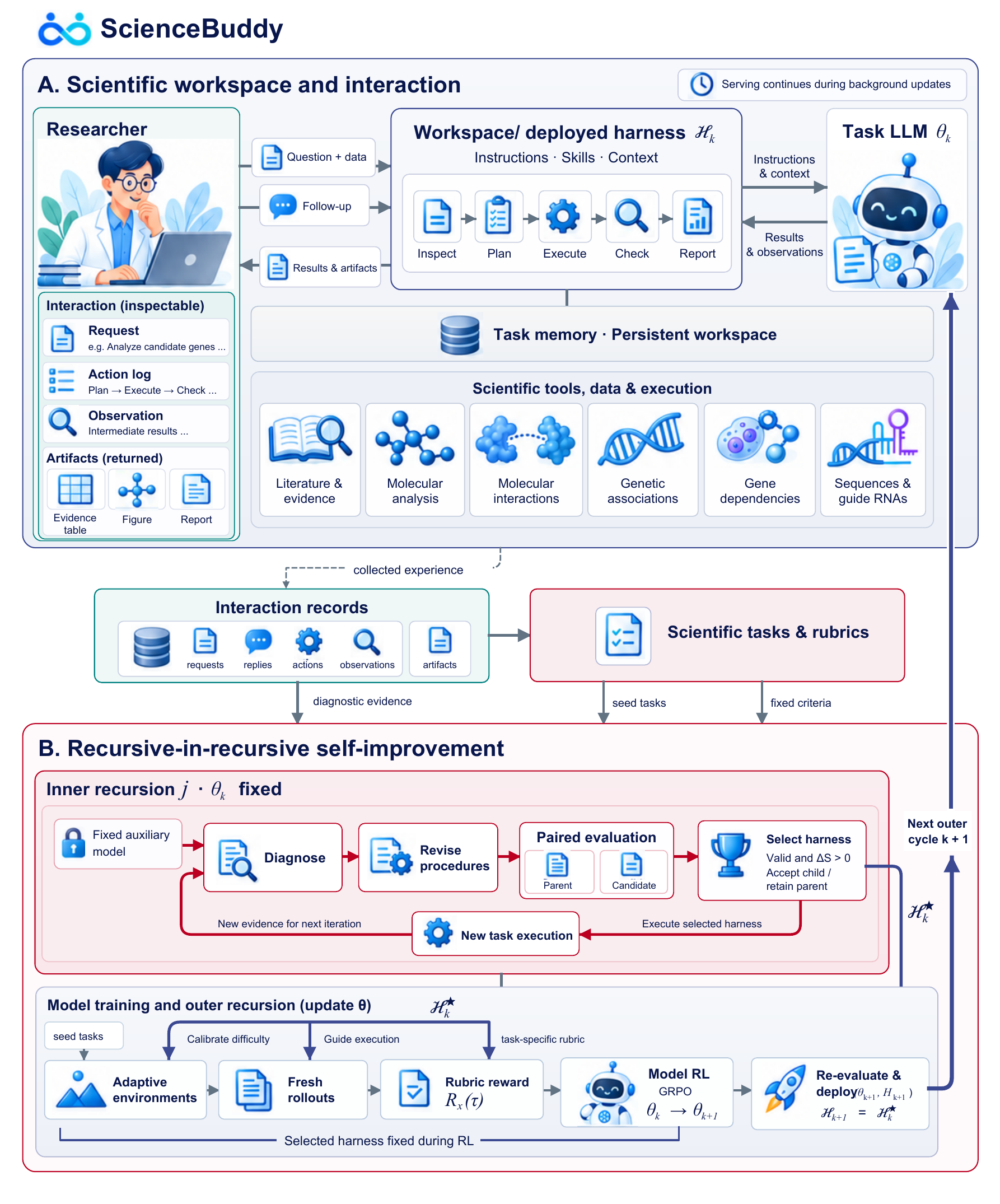}
\caption{\textbf{ScienceBuddy: an interactive scientific workspace with recursive-in-recursive self-improvement.} \textbf{Top: Scientific workspace and researcher interaction.} Researchers submit questions and data, inspect execution traces and artifacts, and refine analyses through follow-up exchanges. The deployed harness organizes the task model's instructions, skills, and context, connecting it to scientific tools, data, and a persistent workspace. \textbf{Middle: From interactions to learning signals.} Requests, replies, actions, observations, and artifacts provide diagnostic evidence and jointly establish executable scientific tasks and task-specific evaluation rubrics. \textbf{Bottom: Coupled harness and model improvement.} In the \textbf{inner recursion}, the task model remains fixed while a fixed auxiliary model diagnoses failures and proposes bounded procedural edits. Valid candidates are accepted only when they improve paired development evaluation; execution under the selected harness supplies evidence for further refinement. In the \textbf{outer recursion}, the selected harness guides environment-difficulty calibration and fresh on-policy rollouts. Rubric rewards drive GRPO updates to the task model, while the harness and evaluation rubrics remain fixed. The updated model and inherited harness are re-evaluated and deployed together, generating researcher interactions for the next cycle. These background updates proceed while the online service remains available.}
\label{fig:sciencebuddy-combined}
\label{fig:sciencebuddy-overview}
\label{fig:recursive-improvement}
\label{fig:coordination}
\end{minipage}%
}
\end{figure}

\clearpage
\setcounter{tocdepth}{4}
\begingroup
\setlength{\parskip}{0pt}
\setlength{\cftbeforesecskip}{3pt}
\setlength{\cftbeforesubsecskip}{0.5pt}
\setlength{\cftbeforesubsubsecskip}{0pt}
\linespread{1}\fontsize{9}{11}\selectfont
\tableofcontents
\endgroup
\clearpage
\section{Introduction}
\label{sec:introduction}

Scientific research proceeds through analysis, inspection, and revision. Language-model agents can assist by retrieving evidence, querying databases, and executing computational workflows \citep{biomni,codeact,labbench}. Researchers then clarify assumptions, question conclusions, and request additional checks. These exchanges reveal how scientific work should be conducted and assessed, but correcting an answer within a conversation does not establish improvement across tasks. This motivates our central question: \emph{How can a scientific agent turn collaboration with researchers into sustained improvements in its working procedures and underlying capabilities?}


Prior work establishes foundations for this problem. Reflection and harness optimization revise reusable instructions and execution procedures \citep{reflexion,gepa,ace,metaharness}; interaction-driven adaptation and rubric-based reinforcement learning provide mechanisms for model improvement \citep{openclaw,seal,rar}. Joint adaptation also has precedent: SIA updates both harnesses and model weights, including for single-cell RNA denoising \citep{sia}, while HELIX connects harness evolution to model-training data construction \citep{helix}. In science, AgentBuild constructs agents from scientist-authored rubrics, curricula, and knowledge bases \citep{agentbuild}. We investigate how \emph{collaboration itself} can supply the tasks and assessment criteria that coordinate repeated procedural and policy learning.

We introduce \textbf{\method{}}, an \textbf{interactive scientific research workspace for continual learning from researcher collaboration}. \Cref{fig:sciencebuddy-combined} provides an overview of the workspace, which combines scientific tools and reference resources \citep{biomni} with data upload, executable analysis, persistent files, and inspectable traces and artifacts. Researchers refine their requests through dialogue, while a pluggable harness organizes model behavior through instructions, reusable skills, and context-management procedures. Separating this editable harness from the scientific infrastructure makes procedural changes explicit and evaluable. Requests, clarifications, execution records, and artifacts jointly establish task objectives, constraints, and success criteria. We consolidate these criteria into task-specific rubrics and package the corresponding instructions, inputs, and environments as executable Harbor tasks \citep{harbor}. Researcher replies inform these criteria without serving as unquestioned correctness labels. The resulting tasks support both procedural diagnosis and evaluation of fresh policy rollouts, using executable checks and fixed judges as appropriate.

On this foundation, we propose \textbf{recursive-in-recursive self-improvement} (\Cref{fig:sciencebuddy-combined}). The \textbf{inner recursion} holds the task model fixed while a separate, fixed auxiliary model diagnoses failures and proposes bounded edits to instructions, skills, or context settings. Candidates are accepted only when they satisfy edit constraints and improve paired development evaluation \citep{skillopt,skillgen}. Further execution supplies evidence for the next revision. The \textbf{outer recursion} calibrates augmented task environments against the current model and selected harness \citep{envevolution}, then trains on fresh on-policy rollouts with task-specific rubric rewards and GRPO \citep{rar,deepseekmath}. The harness and rubrics remain fixed during training; historical interactions provide task definitions and diagnostic evidence rather than on-policy training samples.

The coupling is bidirectional: harness revisions shape training trajectories and task difficulty, while model updates change the effectiveness of inherited procedures. Background improvement proceeds alongside the online service. After re-evaluation, the updated model--harness pair returns to researchers, whose interactions initiate the next cycle. All harness and environment versions are retained for subsequent evolution. Thus, each outer cycle learns through an inner adaptation process and changes the model that participates in the next.

Our case studies examine real researcher interactions, harness revision with a fixed task model, and model learning with a fixed harness. The benchmark cases cover four task families from LAB-Bench and Biomni-Eval1: literature reading, database judgments, protocol troubleshooting, and gene and variant assessment \citep{labbench,biomni}. Holding one component fixed provides a focused view of changes in the other: the harness case measures first-response accuracy on feedback-accessible evaluation tasks, while the model case measures problem coverage on a common panel under H0. These studies connect the proposed framework to observable improvements in scientific task execution.

We release \method{} as an interactive research product, bringing scientific assistance and continual capability improvement into a shared workspace for researchers. This release makes our proposed paradigm available to the scientific community and takes a step toward \textbf{discovery intelligence}, where scientific agents evolve through sustained collaboration with the researchers they support.

\textbf{Contributions.} Our contributions are fourfold:
\begin{itemize}
\item \textbf{A released scientific research workspace.} We develop and release \method{}, an interactive product that helps researchers carry out scientific tasks by connecting researcher dialogue, executable analysis, inspectable artifacts, and a pluggable harness within a persistent workspace.
\item \textbf{Interaction-grounded tasks and supervision.} We formulate a workflow for deriving executable tasks and evaluation rubrics from collaboration, with validated environment augmentation calibrated to current capabilities.
\item \textbf{Recursive-in-recursive self-improvement.} We introduce a paradigm for model--harness co-design that couples evaluated harness evolution with rubric-supervised model reinforcement learning, returning the updated system to researchers for renewed interaction and adaptation.
\item \textbf{Case-study evidence for procedural and model learning.} We examine real researcher interactions, fixed-model harness evolution, and model learning under a fixed harness, relating the proposed framework to improved scientific task execution and broader problem coverage.
\end{itemize}

\clearpage
\section{ScienceBuddy}
\label{sec:method}

\method{} is an interactive scientific research workspace that brings evidence access, computational analysis, and methodological guidance into a single conversational workflow. Researchers can introduce questions together with their data, inspect the resulting analyses, and refine the work through subsequent exchanges. Built on this foundation, \method{} supports \emph{recursive-in-recursive self-improvement}: an inner process revises and evaluates the agent's harness while keeping the task model fixed (\Cref{sec:inner}), and an outer process applies continual reinforcement learning to trajectories generated under the evolving harness (\Cref{sec:outer}). The updated model then returns to further harness adaptation, coupling improvements in working procedures with improvements in the model that executes them (\Cref{sec:coordination}). \Cref{fig:sciencebuddy-combined} summarizes the coupled harness and model improvement process.

\subsection{Scientific Workspace \& Agent Harness}
\label{sec:harness}

We first describe three system components: scientific tools and execution environments, agent execution and researcher interaction, and modular infrastructure with a pluggable harness. \Cref{fig:sciencebuddy-combined} summarizes the scientific workspace and researcher interaction.

\paragraph{Scientific tools and execution environments.}
\method{} provides access to a catalog of 224 tools across 22 functional modules, spanning genomics, molecular and cancer biology, pharmacology, bioimaging, literature retrieval, and database queries. The runtime supports Python, R, and Bash execution, combining scientific libraries with data processing, statistical analysis, and visualization. Online database interfaces and a local data lake provide complementary access to biomedical evidence. Researcher-provided documents, tables, sequences, and images enter a persistent workspace that retains inputs, intermediate files, and generated outputs. Interface and environment details appear in \Cref{sec:dataset_details}. The scientific tool catalog and execution utilities are derived from \citet{biomni}.

\paragraph{Agent execution and researcher interaction.}

We follow a ReAct-style reasoning--action--observation loop \citep{react}, alternating reasoning, code or tool execution, and observation. Researchers submit questions, upload supporting data, and provide follow-up instructions through the Chat view. The Trajectory view presents the chronological execution record, an event timeline, and details of selected events. Compute and Results panels provide access to execution activity and generated artifacts. Conversation history and workspace files preserve task context across exchanges, allowing researchers to inspect the agent's work and request revisions. \Cref{sec:interface} illustrates both interface views.

\paragraph{Modular infrastructure and pluggable harness.}
\method{} separates the agent harness from the infrastructure that manages researcher interactions, task execution, and persistent workspaces. A common execution interface specifies the task context supplied to the harness and the responses and execution records returned to the platform. Alternative agentic harnesses can be integrated by implementing this interface, while sharing the same task-management and storage services. Within this architecture, instructions, skills, and selected context-management procedures constitute the editable components of the harness. Recursive improvement revises these components while keeping the surrounding infrastructure fixed, allowing changes in scientific problem-solving procedures to be evaluated under consistent execution conditions (\Cref{sec:inner}).

\subsection{Interaction Formulation and Learning Signals}
\label{sec:interaction}

\paragraph{Interaction formulation.}
Let $x$ denote a research request and its inputs, $\pi_\theta$ the task model, $H$ the harness, and $h_t=(x,a_0,o_1,\ldots,a_{t-1},o_t)$ the history, with $h_0=(x)$. An action $a_t$ is executable code, a tool call, or a researcher-facing response. The observation $o_{t+1}=(e_{t+1},u_{t+1})$ records environment output or execution status $e_{t+1}$ and an optional researcher reply $u_{t+1}$, with $u_{t+1}=\bot$ when absent. The harness constructs model context $C_H(h_t)$ from history, memory, skills, and tool descriptions. Allowing for a scheduled deterministic action $d_H(h_t)$, such as input inspection, the joint policy and trajectory are
\begin{equation}
\begin{aligned}
\mu_{\theta,H}(a\mid h_t)
&=\begin{cases}
\delta_{d_H(h_t)}(a), & \text{if a harness action is scheduled},\\
\pi_\theta(a\mid C_H(h_t)), & \text{otherwise},
\end{cases}\\
a_t&\sim\mu_{\theta,H}(\cdot\mid h_t),\qquad
\tau=(x,a_0,o_1,\ldots,a_{T-1},o_T).
\end{aligned}
\label{eq:interaction_policy}
\end{equation}
Here $\delta$ denotes a point mass and $T$ counts execution steps. A researcher-facing response may follow several tool steps; a tool observation alone does not constitute a researcher turn or user feedback.

\begin{figure}[htbp]
\centering
\includegraphics[width=\linewidth]{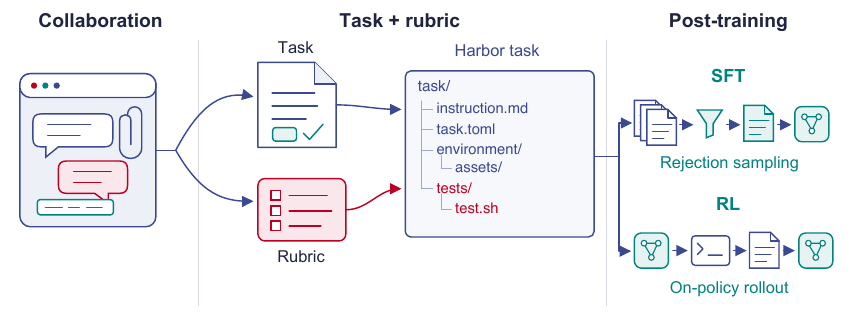}
\caption{\textbf{Collaboration-derived Harbor tasks for post-training.} The collaboration supplies both the task definition and rubric. In the schematic file tree, \texttt{instruction.md} defines the task, \texttt{task.toml} configures execution, \texttt{environment/} holds task assets, and \texttt{tests/test.sh} invokes rubric-based assessment. The resulting tasks support SFT through rejection sampling and RL through on-policy rollouts.}
\label{fig:trajectory-rubric}
\end{figure}

\paragraph{From collaboration to tasks and rubrics.}
The collaboration record supplies two complementary artifacts: a self-contained task and its evaluation rubric (\Cref{fig:trajectory-rubric}). Related turns are consolidated around a scientific objective, with independently solvable objectives separated. The task instruction preserves the final requirements and inputs without importing the historical answer. Unlike earlier task-only packaging followed by expert annotation, the current workflow also derives the rubric from the full collaboration trajectory:
\begin{equation}
\mathcal C(x)=\operatorname{ConstructRubric}
\bigl(\tau_x^{\mathrm{collab}};I_x,A_x\bigr),
\label{eq:trajectory_rubric}
\end{equation}
where $\tau_x^{\mathrm{collab}}$ is the source collaboration, $I_x$ the reconstructed instruction, and $A_x$ the required assets. Criteria cover task scope, methodological requirements, evidence, and expected artifacts. Conflicting requirements are resolved before scoring; historical answers and researcher approval are not automatically treated as scientific ground truth.

\paragraph{Harbor tasks for post-training.}
The task package combines the instruction, input assets, execution environment $\mathcal E_x$, and rubric:
\begin{equation}
\mathcal P_x=\bigl(I_x,A_x,\mathcal E_x,\mathcal C(x)\bigr).
\label{eq:collaboration_task_package}
\end{equation}
Instructions, configuration, assets, and rubric-based tests are organized as Harbor tasks \citep{harbor}. The same tasks support two post-training routes: SFT retains rubric-qualified generated trajectories through rejection sampling, while RL collects fresh on-policy rollouts and uses rubric scores as rewards. Input and runtime checks establish executability; rubric-based checks and a fixed judge assess scientific requirements. The rubric remains fixed within each post-training stage.

\subsection{Inner Recursion: Feedback-guided Harness Improvement}
\label{sec:inner}

At inner step $j$ of outer cycle $k$, the active harness $H_{k,j}$ is the \emph{parent}, and its proposed revision $\widetilde H_{k,j+1}$ is a \emph{candidate}. An accepted candidate becomes the \emph{child} $H_{k,j+1}$. Otherwise, the parent remains active.

\paragraph{Feedback-guided diagnosis.}
Within outer cycle $k$, the task-model parameters $\theta_k$ remain fixed. We use GPT-6 Astra as a separate, fixed auxiliary model for trajectory diagnosis and harness editing. It reviews recent trajectories and rubric evaluations, identifies unmet criteria, and cites the relevant actions and observations. Following evidence-based trajectory diagnosis \citep{agentrx}, it maps these findings to a candidate procedural edit. Task-specific answers and newly supplied facts remain local to the task.

\paragraph{Harness revision.}
Let $E_{k,j}$ contain the selected trajectories, rubric feedback, and edit history for harness $H_{k,j}$. The auxiliary model proposes a bounded update,
\begin{equation}
\widetilde H_{k,j+1}=U(H_{k,j},E_{k,j};\theta_k).
\label{eq:harness_update}
\end{equation}
Each proposal adds, removes, or revises one scoped skill, edits an instruction, or changes one exposed context setting, leaving other components unchanged \citep{apsf,skillopt}. A schema check enforces the permitted edit scope and size budget. Tools, execution infrastructure, rubrics, and evaluators remain fixed. This is a procedural update; neither the task model nor the auxiliary model receives gradient updates.

\paragraph{Evaluation and recursive refinement.}
Parent and candidate are evaluated on identical development tasks, seeds, and execution budgets using frozen task rubrics. Let $\bar S_k(H)$ be the mean normalized rubric score and $\mathrm{Valid}(H)$ indicate compliance with the edit constraints. Write $\Delta_{k,j}=\bar S_k(\widetilde H_{k,j+1})-\bar S_k(H_{k,j})$. The proposed acceptance rule is
\begin{equation}
H_{k,j+1}=\begin{cases}
\widetilde H_{k,j+1}, & \mathrm{Valid}(\widetilde H_{k,j+1})\ \land\ \Delta_{k,j}>0,\\
H_{k,j}, & \text{otherwise}.
\end{cases}
\label{eq:harness_selection}
\end{equation}
Evaluation includes previously successful tasks to account for regressions \citep{skillgen}, and ties retain the parent. Rejected edits and score changes remain in the optimizer's history. The selected harness then executes new training tasks, whose trajectories supply evidence for the next revision. Iteration continues until the proposal budget or outer collection boundary is reached. Development tasks are separate from policy-training tasks and the final held-out test set, which never informs editing or selection.

\subsection{Outer Recursion: Continual Model Reinforcement Learning}
\label{sec:outer}
\label{sec:learning}

%
\paragraph{Environment augmentation under an evolving harness.}
As the harness evolves, previously challenging tasks may become routine, reducing their value for further model training. We therefore calibrate environment difficulty through pilot execution with the current task model and selected harness. Following environment evolution \citep{envevolution}, we augment researcher-derived tasks by varying scientific inputs and analysis conditions or extending dependencies between computational steps. The validated environments then supply fresh RL rollouts.

\paragraph{Task-adaptive rubric rewards.}
For each task $x$, a fixed rubric composer derives task-specific criteria from the source collaboration trajectory and its reconstructed objective, inputs, and required outputs (\Cref{sec:interaction}), following task-adaptive rubric construction \citep{adarubric}. The resulting rubric $\mathcal C(x)$ combines task-specific correctness checks with relevant evidence and artifact requirements. Each criterion has a nonnegative importance weight $w_c(x)$, assigned before rollout evaluation, and a satisfaction score $v_c(x,\tau)\in[0,1]$. We use executable checks where available and a fixed judge for criteria requiring scientific interpretation \citep{robustrubrics}. Following rubric-based reward aggregation \citep{rar}, the trajectory reward is
\begin{equation}
R_x(\tau)=
\frac{\sum_{c\in\mathcal C(x)}w_c(x)\,v_c(x,\tau)}
{\sum_{c\in\mathcal C(x)}w_c(x)},
\qquad \sum_{c\in\mathcal C(x)}w_c(x)>0.
\label{eq:reward}
\end{equation}
Rubrics vary across tasks but remain fixed during optimization and paired harness evaluation. The terminal reward supplies a trajectory-level advantage shared across generated tokens. We use GRPO \citep{deepseekmath}; its objective and implementation details are given in \Cref{sec:grpo_details}.

\paragraph{Model updates and renewed harness adaptation.}
At outer cycle $k$, we maximize the expected trajectory reward under the selected harness:
\begin{equation}
\max_{\theta}J_k(\theta),\qquad
J_k(\theta)=
\mathbb E_{x\sim q_k}
\mathbb E_{\tau\sim\pi_{\theta,H_k^\star}(\cdot\mid x)}
\bigl[R_x(\tau)\bigr].
\label{eq:rl_objective}
\end{equation}
Here $q_k$ is the training-task distribution over the validated seed environments and augmented variants at outer cycle $k$, and $\pi_{\theta,H_k^\star}$ is the trajectory distribution induced by the task model under the fixed harness $H_k^\star$. The GRPO update yields $\theta_{k+1}$. Because harness effectiveness depends on its interaction with the task model \citep{metaharness}, we re-evaluate the selected harness under the updated model before deploying $(\theta_{k+1},H_{k+1})$, with $H_{k+1}=H_k^\star$. Researcher interactions with this pair provide evidence for the next inner-adaptation phase and outer update cycle (\Cref{sec:coordination}).

\subsection{Coordinating Recursive-in-Recursive Improvement}
\label{sec:coordination}

\paragraph{Nested update schedule.}
\method{} serves researchers with model $\theta_k$ and harness $H_k$ over a fixed collection interval. The resulting interactions and feedback initiate a background update cycle, asynchronous with the online service: harness improvement proceeds with $\theta_k$ fixed, followed by model RL under the selected harness. After re-evaluation, the updated model--harness pair is deployed to support increasingly demanding research tasks. Subsequent researcher interactions provide the evidence for the next cycle (\Cref{alg:rsi}).

\paragraph{Cross-cycle experience and re-evaluation.}
All harness versions and task environments are retained for subsequent evolution. Inherited harnesses are re-evaluated under the updated task model before deployment or reuse.

\begin{algorithm}[H]
\caption{Recursive-in-recursive improvement with asynchronous online service}
\label{alg:rsi}
\footnotesize
\begin{algorithmic}[1]
\Require Initial $(\theta_0,H_0)$, collection interval $\Delta$, training environments $\mathcal T$,
\Statex \hspace{\algorithmicindent} development tasks, inner budgets $J_k$, RL budgets, and outer count $K$
\State Initialize evidence buffer $\mathcal B$ and edit history $\mathcal L$; deploy $(\theta_0,H_0)$
\For{$k=0,\ldots,K-1$}
  \State $E_k\gets\operatorname{Collect}_{\Delta}(\theta_k,H_k)$; $\mathcal B\gets\mathcal B\cup E_k$
  \Statex \hspace{\algorithmicindent}\emph{Background updates; the online service continues with $(\theta_k,H_k)$.}
  \State $H_{k,0}\gets H_k$; evaluate $\bar S_k(H_{k,0})$; $j\gets0$
  \While{$j<J_k$ and the inner execution budget remains}
    \State Append fresh task evidence under $(\theta_k,H_{k,j})$ to $\mathcal B$
    \State $E_{k,j}\gets\operatorname{Read}(\mathcal B,\mathcal L;H_{k,j})$
    \State $\widetilde H_{k,j+1}\gets U(H_{k,j},E_{k,j};\theta_k)$ \Comment{\Cref{sec:inner}}
    \State Validate and, if valid, evaluate $\widetilde H_{k,j+1}$ under paired development conditions
    \State Select $H_{k,j+1}$ by \Cref{eq:harness_selection}; record the decision in $\mathcal L$
    \State $j\gets j+1$
  \EndWhile
  \State $H_k^\star\gets H_{k,j}$; $\mathcal T_k\gets\operatorname{Augment}(\mathcal T;\theta_k,H_k^\star)$
  \State $\theta_{k+1}\gets\operatorname{RLUpdate}(\theta_k;H_k^\star,\mathcal T_k,R)$ \Comment{\Cref{sec:outer}}
  \Statex \hspace{\algorithmicindent}Use fresh batches $\mathcal D_{k,t}$, \Cref{eq:reward,eq:grpo}; retire batches after optimization.
  \State $H_{k+1}\gets H_k^\star$; re-evaluate $(\theta_{k+1},H_{k+1})$
  \State Retain all harness and environment versions; $\mathcal T\gets\mathcal T\cup\mathcal T_k$
  \State Deploy $(\theta_{k+1},H_{k+1})$
\EndFor
\State \Return $(\theta_K,H_K)$
\end{algorithmic}
\end{algorithm}
\FloatBarrier

\Needspace{8\baselineskip}
\section{Scientific Workspace and User Experience}
\label{sec:workspace}

\paragraph{Scientific scope.}
\method{} combines multimodal input, long-context agentic reasoning, and researcher interaction within a shared scientific workspace. Its document handling and execution interfaces support multiple scientific domains, while the current tools and data specialize in biomedicine. The following recorded session illustrates how researchers connect visual scientific material to target analysis, evidence retrieval, and further questions.

\begin{figure}[H]
\centering
\includegraphics[width=\linewidth]{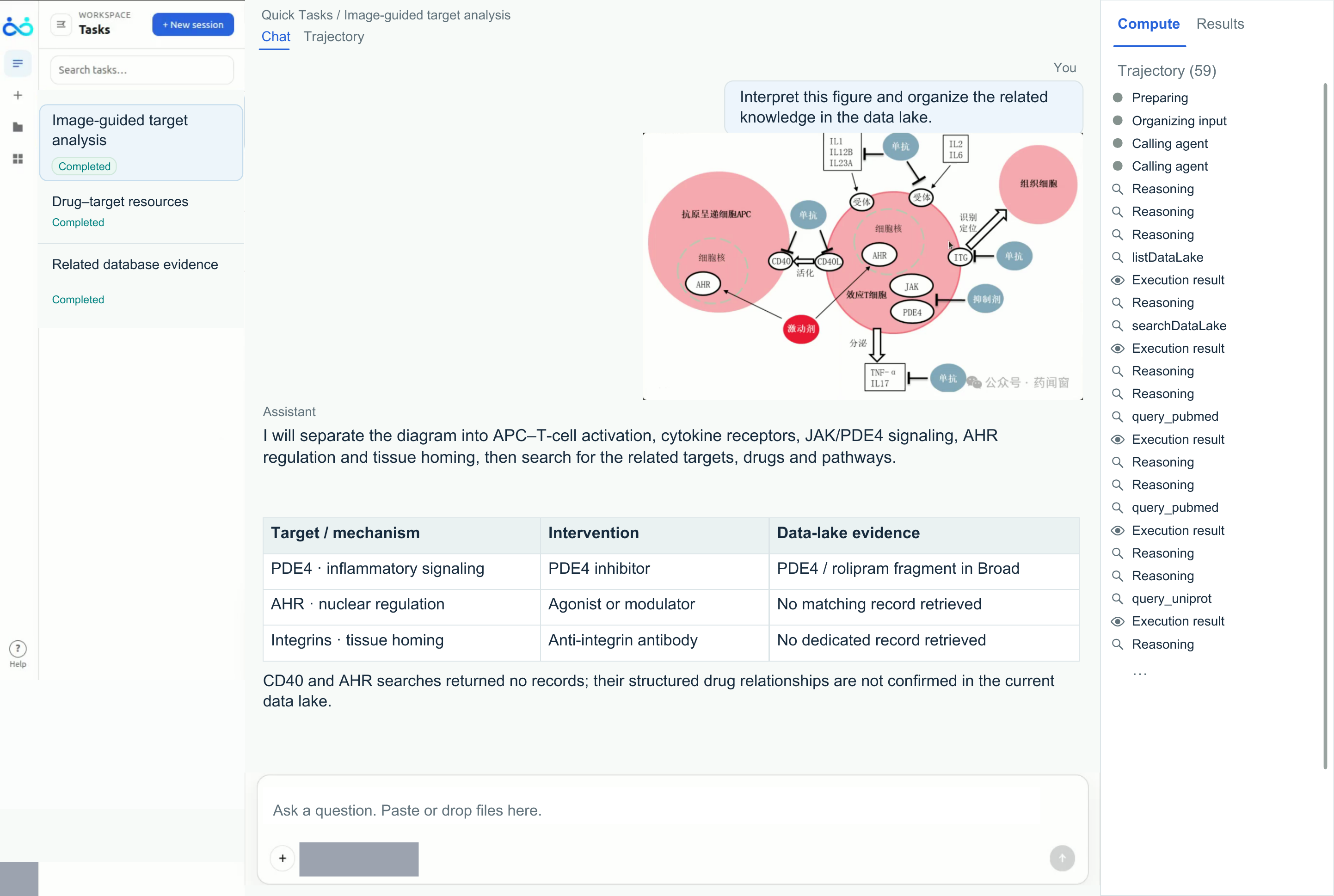}
\caption{\textbf{A workspace for multimodal scientific analysis.} A researcher supplies a scientific diagram and requests related knowledge. The Chat view connects visual interpretation to a structured target--evidence table, while the Compute panel exposes execution records. Retrieved evidence and gaps in the available data remain visible for researcher inspection. UI text and dialogue are reconstructed in English from the recording; uploaded figures retain their original appearance and language. Account and model identifiers are masked.}
\label{fig:workspace-demo-chat}
\end{figure}

\paragraph{Multimodal input and evidence inspection.}
Researchers can supply documents, tables, biological sequences, and images alongside natural-language requests. In \Cref{fig:workspace-demo-chat}, an uploaded immune-signaling diagram guides the identification of molecular targets and the organization of related drug and pathway knowledge. The response connects visual entities to an evidence table, distinguishing a retrieved PDE4/rolipram fragment from CD40 and AHR searches that returned no matches. The conversation, input composer, and Compute panel bring the scientific material, response, and execution history into one inspectable view. Original interface captures appear in \Cref{sec:interface}.

\begin{figure}[p]
\centering
\includegraphics[width=\linewidth,height=0.83\textheight,keepaspectratio]{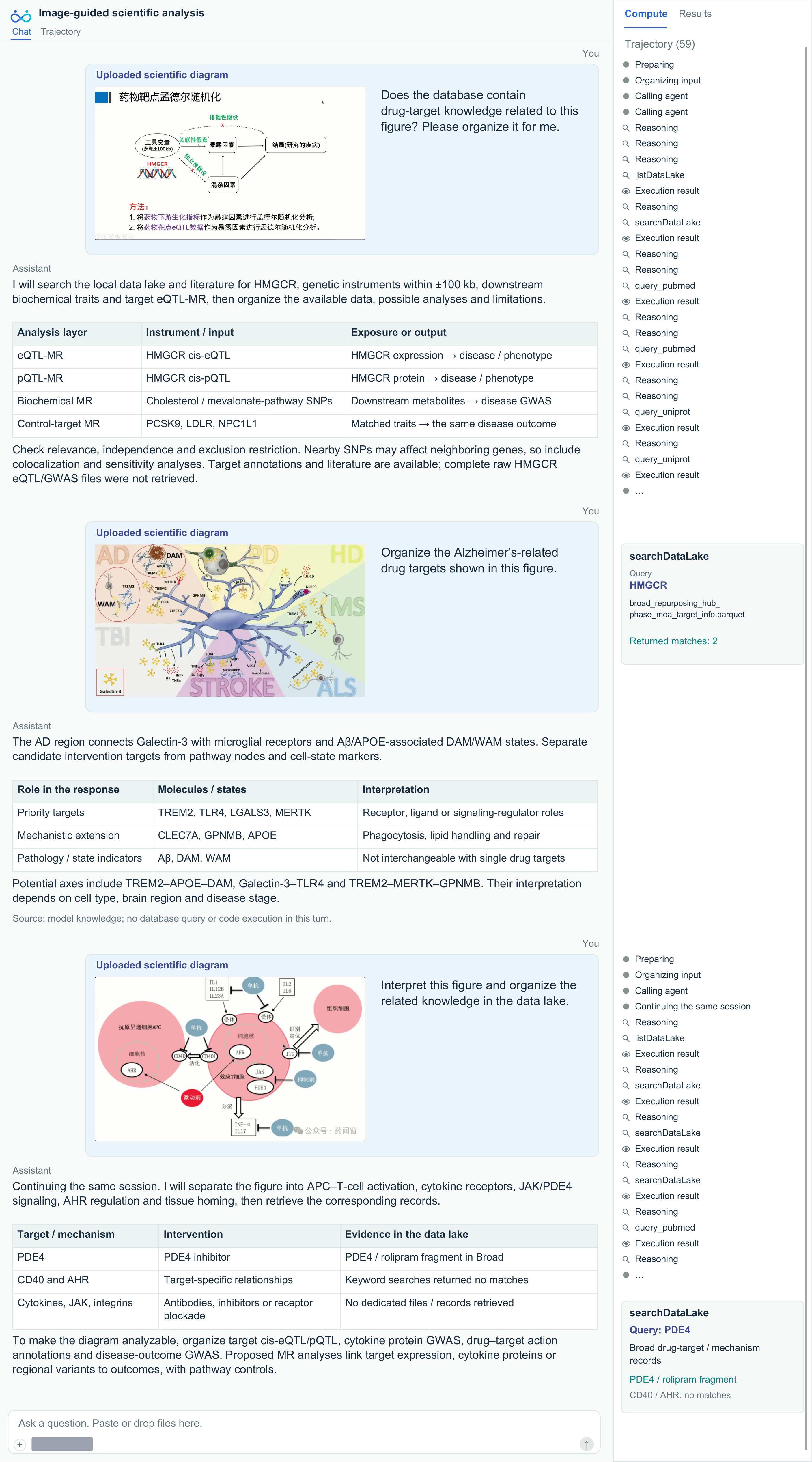}
\caption{\textbf{Multimodal input, long-context agentic reasoning, and researcher interaction.} Three successive image-based requests direct target analysis across a continuing scientific session. Uploaded diagrams, assistant interpretations, and evidence tables are paired with the recorded execution history, showing how researcher direction and retained context connect successive rounds of work. Proposed analyses are not executed experiments. English dialogue and UI are reconstructed from recorded moments; uploaded figures retain their original language, omitted events are marked, and identifiers are masked.}
\label{fig:workspace-demo-inspection}
\end{figure}

\paragraph{Long-context agentic reasoning.}
\Cref{fig:workspace-demo-inspection} follows three image-based requests in a continuing session: an HMGCR Mendelian-randomization diagram, an Alzheimer's-related microglial network, and an immune-signaling diagram. The agent interprets each image through reasoning, retrieval, and synthesis; the later execution explicitly resumes the same session with prior exchanges available. The trace records repeated data-lake searches and literature/protein queries; the middle response instead uses model knowledge without a new database query.

\paragraph{Researcher interaction.}
The researcher directs the work by introducing new diagrams, changing the scientific focus, and explicitly requesting database evidence. Successive responses organize targets, distinguish pathways from cell-state markers, and identify data needed for further analysis. Retained dialogue and evidence support subsequent requests and the derivation of task objectives and evaluation criteria (\Cref{sec:interaction}).

\FloatBarrier

\begin{figure}[H]
\centering
\includegraphics[width=\linewidth]{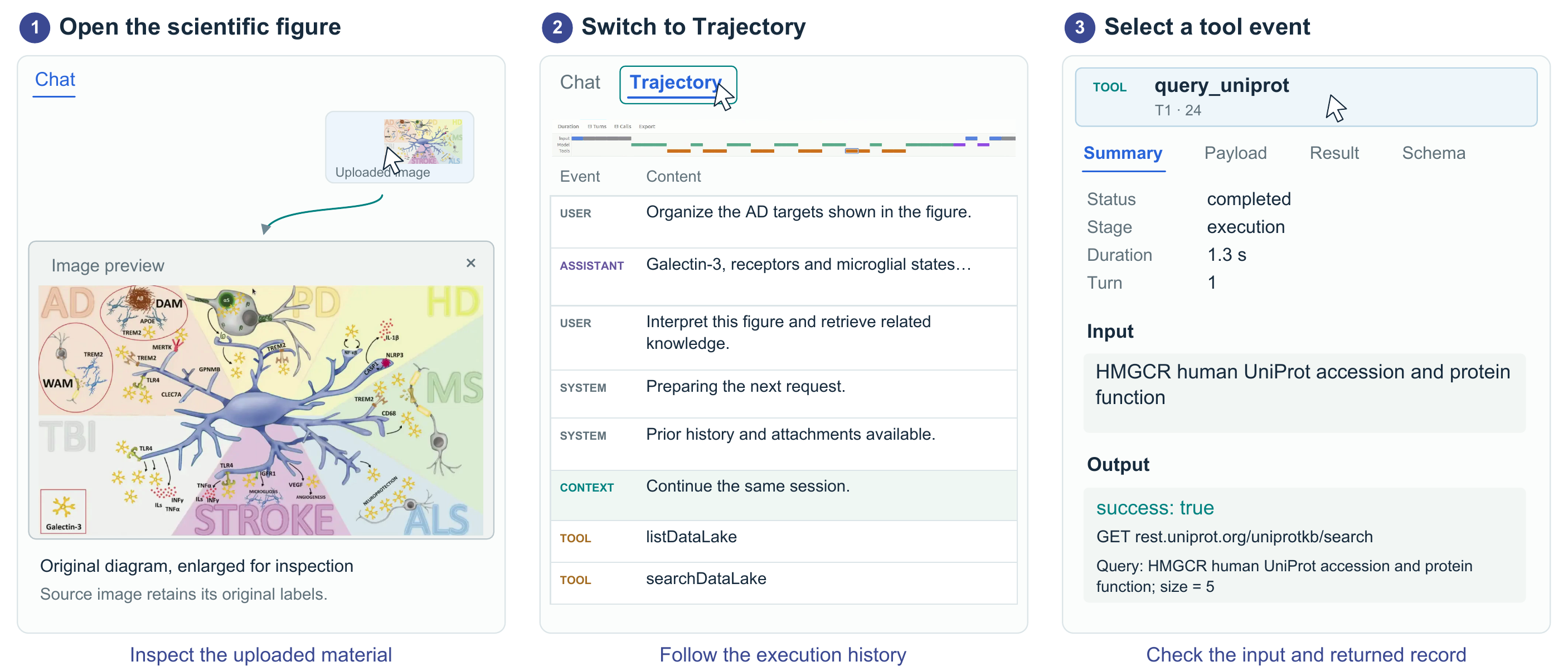}
\caption{\textbf{Researcher inspection beyond the conversation.} Opening an uploaded image reveals its scientific details; switching to Trajectory exposes the execution history; selecting a tool event opens its input, output, and metadata. The example revisits an earlier HMGCR protein lookup within the continuing session. English interface reconstructions highlight controls used in the recording; the uploaded diagram and timeline retain source pixels. Cursor markers indicate the inspected controls.}
\label{fig:researcher_inspection_controls}
\end{figure}

\paragraph{Researcher inspection through interface controls.}
Researcher interaction also includes navigation and inspection actions beyond conversational input (\Cref{fig:researcher_inspection_controls}). In the demonstration, the researcher opens an uploaded diagram at a larger scale, switches from Chat to Trajectory, and selects a tool event to inspect its metadata, input, and output. The selected UniProt event exposes an earlier HMGCR lookup while later requests remain in the same session. These controls let the researcher examine source material, follow the execution history, and revisit the basis of a response without starting a new conversation.

\FloatBarrier

\Needspace{8\baselineskip}
\section{Case Studies}
\label{sec:experiments}
\label{sec:results}
\label{sec:case_study}

We present four distinct case studies of \method{}'s scientific assistance and self-improvement. Each addresses a separate research question:

\begin{list}{}{\setlength{\leftmargin}{0pt}\setlength{\itemindent}{0pt}\setlength{\itemsep}{3pt}\setlength{\parsep}{0pt}\setlength{\topsep}{3pt}}
\item \textbf{RQ1: Researcher interaction.} How does researcher feedback guide scientific assistance and reveal task objectives and evaluation criteria? (\Cref{sec:interaction_cases})
\item \textbf{RQ2: Coupled Recursive-in-Recursive improvement.} Can alternating harness refinement and model learning sustain improvement across cycles and broaden scientific task performance? (\Cref{sec:three_cycle_dynamics})
\item \textbf{RQ3: Harness adaptation.} Can harness adaptation improve scientific task performance without changing model weights? (\Cref{sec:feedback_results})
\item \textbf{RQ4: Model learning.} Can reinforcement learning expand scientific problem-solving capability under a fixed harness? (\Cref{sec:recursive_dynamics})
\end{list}

\subsection{ScienceBuddy Interaction}
\label{sec:interaction_cases}

\paragraph{Setup.}
We examine two real researcher interactions with deployed \method{}. Requests, supplied materials, agent responses, and subsequent researcher input support a qualitative assessment of scientific assistance and opportunities for reinforcement learning (RL) task construction. Readers interested in the concrete researcher wording can consult \Cref{fig:researcher_requests} in the appendix.

\paragraph{Refining a JAK1 investigation.}
A researcher asked \method{} to design a study of JAK1, immunotherapy outcomes, and the immune microenvironment in small-cell lung cancer using public single-cell transcriptomes and IMpower133 bulk RNA data. In response to the scope refinement (\Cref{fig:researcher_requests}a), \method{} organized a gene-specific plan with treatment-by-JAK1 interaction tests, patient-level expression summaries within cell types, and immune-state signatures. The plan assigned Seurat/Scanpy to single-cell analysis, UCell/AUCell to signature scoring, and CellChat/NicheNet to subsequent cell-communication analyses. This plan distinguished treatment-effect modification from prognosis and prioritized mechanistic follow-up.

\paragraph{Connecting evidence in an ARL4C study.}
A researcher requested a presentation connecting the background and results of an ARL4C study, then specified panel selection, conclusions, mechanism schematics, and speaker notes (\Cref{fig:researcher_requests}b). Using text and figure captions organized through Python/PyPDF2, \method{} linked candidate screening to cellular and molecular evidence. It highlighted depletion and conditional knockout comparisons for cellular attribution, blockade for functional dependence, and kinetic and rescue assays for molecular interpretation. Panel-selection rationales and notes linked each scientific claim to its supporting comparison.

\begin{figure}[H]
\centering
\includegraphics[width=0.85\linewidth]{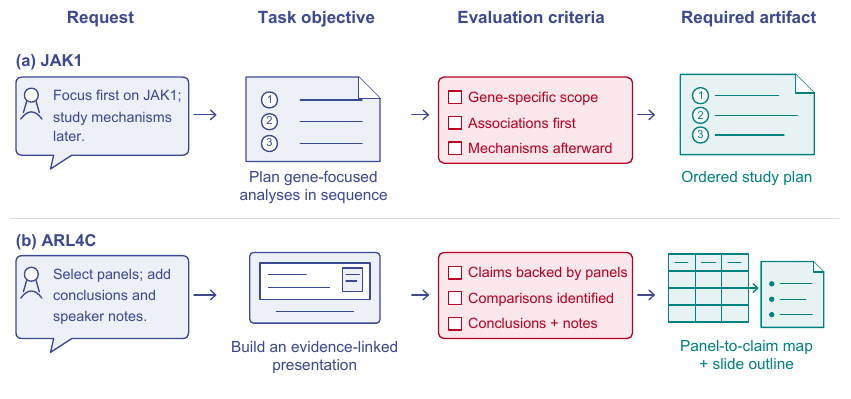}
\caption{\textbf{From researcher requests to task specifications.} (a) A JAK1 scope refinement defines an ordered, gene-focused study plan. (b) ARL4C presentation requirements define an evidence-linked presentation and panel-to-claim map. Requests are translated and abridged from real interactions; the task objectives, evaluation criteria, and required artifacts are illustrative derivations, not archived rubric packages or scored outputs.}
\label{fig:researcher_feedback}
\end{figure}

\paragraph{From requests to task specifications.}
These cases illustrate how researcher requirements translate into task objectives, evaluation criteria, and required artifacts (\Cref{fig:researcher_feedback}). The JAK1 refinement yields a study-planning objective whose criteria preserve gene-specific scope and place association analyses before mechanistic follow-up. The ARL4C request yields a presentation objective whose criteria link claims to supporting panels and comparisons, with conclusions and speaker notes accompanying the slide outline. Such task specifications provide the basis for the trajectory-derived rubrics and post-training tasks described in \Cref{sec:interaction}.

\Needspace{8\baselineskip}
\subsection{Two-Cycle Recursive-in-Recursive Dynamics}
\label{sec:three_cycle_dynamics}

\paragraph{Setup.}
Starting from Qwen3.5-4B and an initial scientific-agent harness, we run three successive co-evolution cycles, indexed by $k=0,1,2$. In cycle $k$, harness refinement starts from $(\theta_k,H_k)$, keeps the model fixed, and performs 10 search steps to select $H_k^\star$ by validation accuracy. Model learning then performs 20 RL updates under the selected harness. The final checkpoint $\theta_{k+1}$ and selected harness $H_{k+1}=H_k^\star$ are carried into the next cycle, where inherited harnesses are reassessed under the updated model. This repeated exchange allows improvements in the model and harness to carry forward, supporting continued system improvement across successive cycles. Dataset and environment details are provided in Appendix~\ref{sec:dataset_details}; detailed experimental settings are deferred to the appendix.

\begin{figure}[htbp]
\centering
\IfFileExists{figures/fig_main.pdf}{%
  \includegraphics[width=\linewidth]{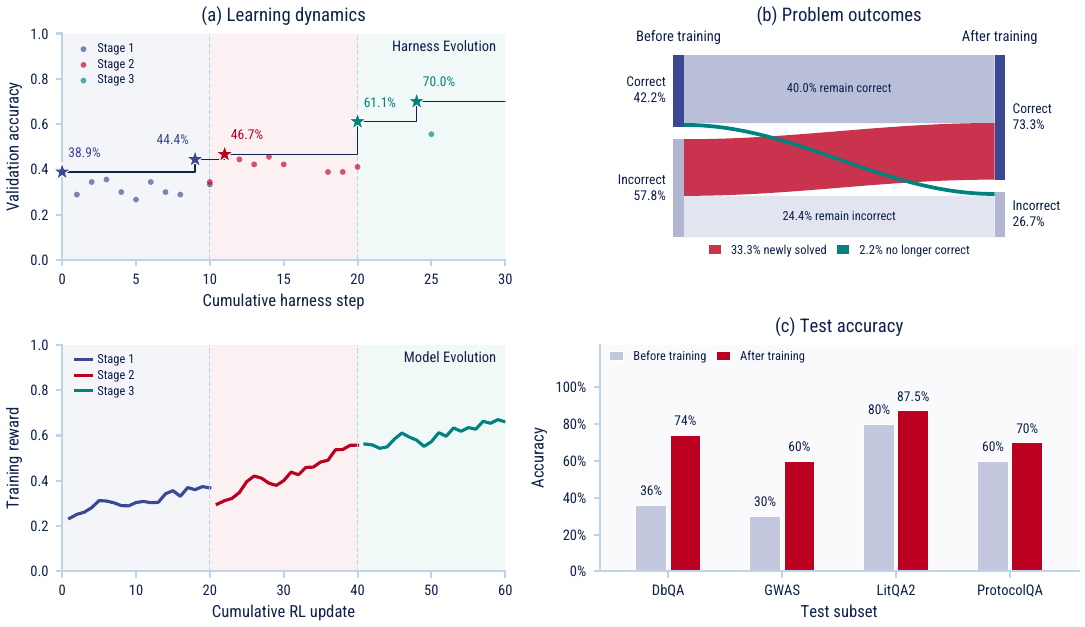}%
}{%
  \fbox{\begin{minipage}[c][58mm][c]{0.94\linewidth}
    \centering
    \textbf{Three-cycle experiment: final figure pending}\par\medskip
    (a) Learning dynamics across three stages\par\smallskip
    Harness validation accuracy and RL training reward\par\medskip
    (b) Paired problem outcomes\quad (c) Test accuracy by subset\par\medskip
    \small Replace with \texttt{rinr-three-cycle-dynamics.pdf}
  \end{minipage}}%
}
\caption{\textbf{Learning dynamics and evaluation across three RinR cycles.}
Each cycle comprises ten harness-evolution steps followed by twenty RL updates; colors identify cycles.
(a) Circles show measured harness validation scores, including rejected candidates. Stars and annotations identify new historical bests.
(b) Outcome transitions pair the initial and final systems on the same test sets.
(c) Test accuracy by scientific task family.
Harness selection uses a separate, fixed validation set.}
\label{fig:three_cycle_dynamics}
\end{figure}

\paragraph{Learning dynamics across cycles.}
Figure~\ref{fig:three_cycle_dynamics}(a) shows consistent improvements within each of the three cycles. Harness refinement increases validation accuracy from 38.9\% to 44.4\%, 34.4\% to 46.7\%, and 61.1\% to 70.0\% in the first, second, and third cycles, respectively. Over the same cycles, mean training reward rises from 33.3\% to 38.8\%, 44.1\% to 60.5\%, and 57.8\% to 69.8\% between the first and second halves of each RL phase.
These gains show that both harness refinement and model training
continue to improve their respective metrics over repeated cycles.

\paragraph{Scientific task performance.}
Figures~\ref{fig:three_cycle_dynamics}(b,c) summarize the improvement in held-out scientific task performance. Overall single-attempt test accuracy increases from 42.2\% to 73.3\%. Among all test problems, 33.3\% transition from incorrect to correct, whereas 2.2\% transition from correct to incorrect. The subset comparison shows gains across all four task families. These results indicate that improvement extends to previously unsolved problems, broadening the system's scientific problem-solving capability.

The next two case studies evaluate harness adaptation and model learning independently, holding model weights or the harness fixed, respectively (\Cref{sec:feedback_results,sec:recursive_dynamics}).

\FloatBarrier

\Needspace{8\baselineskip}
\subsection{Harness Adaptation with a Fixed Model}
\label{sec:feedback_results}

\paragraph{Setup.} We refine and select the harness on an \emph{adaptation set}, then compare the selected and initial harnesses on a separate \emph{validation set}. Tasks from LAB-Bench and Biomni-Eval1 \citep{labbench,biomni} cover literature reading, database judgments, protocol troubleshooting, and gene and variant assessment. Implementation details appear in \Cref{sec:harness_case_protocol}.

\begin{figure}[htbp]
\centering
\includegraphics[width=\linewidth]{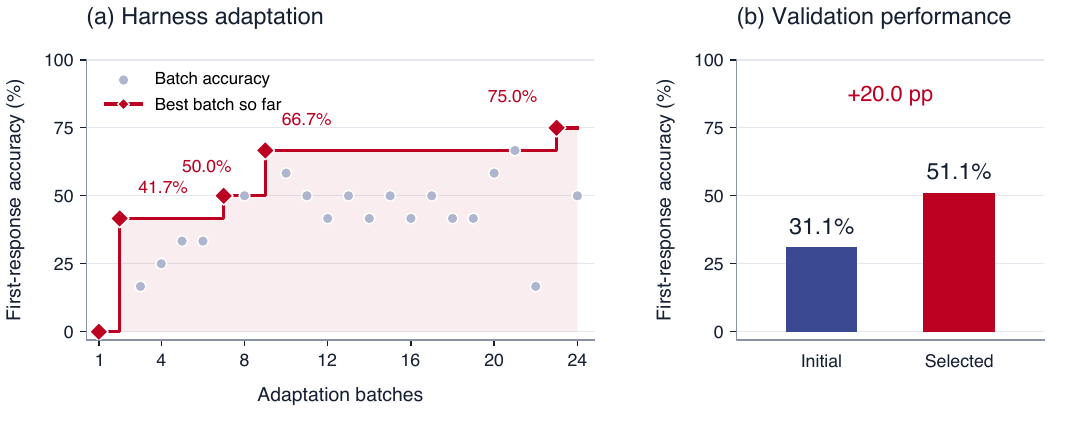}
\caption{\textbf{Harness adaptation and validation performance at fixed model weights.}
(a) First-response accuracy across adaptation batches; the step curve tracks the best batch accuracy observed so far. Batches contain different tasks.
(b) Validation accuracy of the initial harness and the harness selected on the adaptation set: 31.1\% versus 51.1\%, a gain of 20 percentage points. The validation set is used for this comparison, not harness selection.}
\label{fig:harness_training_curves}
\end{figure}

\paragraph{Adaptation and validation performance.}
\Cref{fig:harness_training_curves}a tracks first-response accuracy during harness adaptation: the fraction of tasks answered correctly on the first submission. Across 24 adaptation batches, the best observed batch accuracy reaches 75.0\%. The selected harness is then evaluated on validation tasks, alongside the initial harness (\Cref{fig:harness_training_curves}b). Validation accuracy increases from 31.1\% to 51.1\%, a gain of 20 percentage points with model weights fixed. This improvement demonstrates the effectiveness of revising the agent's working procedures beyond the tasks used for adaptation and selection.

\paragraph{Learned procedures.}
We inspect the selected harness to characterize the procedures retained from interaction. Its four instruction entries and nine scoped skills address Python execution, resource and schema inspection, bounded record lookup, and explicit answer submission. Task-specific procedures include gene-set membership checks, cytoband lookup, and database-specific evidence extraction. These procedures guide the agent in locating and checking scientific records, turning interaction evidence into reusable guidance for task execution. \Cref{sec:harness_case_protocol} describes the revisions and the limits of attributing gains to individual edits or feedback sources.

\FloatBarrier
\Needspace{12\baselineskip}

\subsection{Model Learning with a Fixed Harness}
\label{sec:recursive_dynamics}
\label{sec:core_ablation}

\paragraph{Setup.} We keep the initial harness fixed throughout training and compare the model before and after RL under the same evaluation budget. The learning algorithm and evaluation protocol appear in \Cref{sec:rl_case_protocol}.

\begin{figure}[htbp]
\centering
\includegraphics[width=\linewidth]{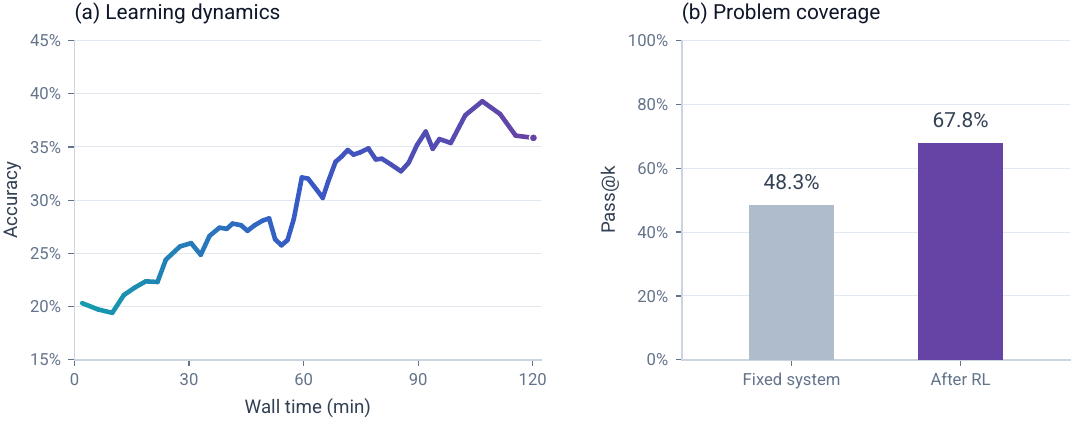}
\caption{\textbf{Model learning and problem coverage under a fixed harness.}
(a) Training accuracy against elapsed time during model learning.
(b) Problem coverage before and after RL, measured by pass@4 under the fixed harness and the same attempt budget. Coverage increases from 48.3\% to 67.8\%, indicating successful solutions to more distinct problems within the same attempt budget.}
\label{fig:learning_coverage}
\end{figure}

\paragraph{Learning dynamics and problem coverage.}
Training accuracy trends upward over approximately two hours of RL (\Cref{fig:learning_coverage}a). To assess whether learning also expands the range of solvable problems, we measure \emph{problem coverage}: the fraction of test problems solved at least once within four attempts. Coverage increases from 48.3\% before RL to 67.8\% afterward (\Cref{fig:learning_coverage}b), a gain of 19.5 percentage points. With both the harness and attempt budget unchanged, the model solves a broader set of scientific problems, demonstrating the effectiveness of model learning as a distinct improvement mechanism.

\FloatBarrier

\section{Related Work}
\label{sec:related}

\textbf{Persistent experience in agents.} Reflexion retains verbal lessons in episodic memory, GEPA searches over prompts using trajectory reflection, and ACE incrementally maintains contextual playbooks \citep{reflexion,gepa,ace}. Meta-Harness extends search to harness code using prior candidates and execution records, while PILOT learns reusable procedures during live execution \citep{metaharness,pilot}. These methods provide mechanisms for persistent procedural adaptation. ScienceBuddy studies how this adaptation generates experience for a second, model-level recursion.

\textbf{Recursive self-improvement.} The Darwin Godel Machine evolves an archive of agents, and Hyperagents makes the meta-level modification procedure part of the editable program \citep{dgm,hyperagents}. SEAL generates data and update directives for parameter adaptation \citep{seal}. ScienceBuddy instead studies a nested dependency between repeated harness adaptation and repeated task-model learning. Its reflector remains fixed, so improved task performance does not imply that the improvement mechanism itself has become stronger.

\textbf{Learning from interaction.} OpenClaw-RL extracts evaluative and directive signals from the states following agent actions, including user replies \citep{openclaw}. RLAnything jointly adapts environments, policies, and reward models \citep{rlanything}. ScienceBuddy studies how these learning processes interact with an evolving harness. User feedback guides procedural revision, while task verification supervises policy trajectories generated under the active harness. The learned model then returns to the next inner process, changing the conditions for further procedural adaptation.


\section{Conclusion}
\label{sec:conclusion}

ScienceBuddy provides an interactive scientific workspace in which researcher collaboration can inform both working procedures and model learning. Its recursive-in-recursive framework connects evaluated harness refinement with rubric-supervised model updates, returning the updated system to further scientific interaction. The case studies illustrate the complementary contributions of these components: researcher requests and follow-up requirements define scientific tasks and assessment criteria; harness revision improves first-response accuracy with the task model fixed; and model learning expands problem coverage under the initial harness. These findings support the framework's procedural and model-learning mechanisms and provide a basis for studying their coordination across continued researcher collaboration.

\FloatBarrier
\phantomsection
\addcontentsline{toc}{section}{References}
\setlength{\bibsep}{1.5pt}
\bibliography{refs}
\clearpage

\supplementary
\section{Implementation Details}
\label{sec:implementation}

This appendix connects the case studies to the scientific workspace and recursive-in-recursive framework described in the main text. We distinguish the fixed-model harness-evolution run from the model-learning case under a fixed harness. The role specifications describe the information boundaries and procedural responsibilities of these components; the policy objective shows how model learning fits within the full recursive procedure.

\subsection{Datasets and Environments}
\label{sec:dataset_details}

\textbf{Task composition.} The 895-task collection contains 96 LitQA2, 511 DbQA, 108 ProtocolQA, and 180 GWAS tasks. LitQA2 and ProtocolQA each have one subtopic, DbQA has ten, and GWAS has four, for 16 subtopics in total. \Cref{tab:dataset_subtopics} reports the number of tasks in each subtopic and the totals for each task family.
\begin{table}[htbp]
\caption{Task counts across four scientific task families and 16 subtopics, totaling 895 tasks.}
\label{tab:dataset_subtopics}
\centering\footnotesize
\begin{tabularx}{\linewidth}{@{}lYr@{}}
\toprule
Family & Subtopic & Count \\
\midrule
LitQA2 & Scientific literature reading & 96 \\
\midrule
DbQA & Disease--gene associations & 39 \\
 & Gene location & 40 \\
 & miRNA targets & 40 \\
 & Mouse tumor gene sets & 80 \\
 & Oncogenic signatures & 40 \\
 & Transcription-factor binding (GTRD) & 40 \\
 & Variant annotation: single sequence & 80 \\
 & Variant annotation: multiple sequences & 72 \\
 & Vaccine-response gene sets & 40 \\
 & Viral protein interactions & 40 \\
 & \textit{Subtotal} & 511 \\
\midrule
ProtocolQA & Experimental protocol troubleshooting & 108 \\
\midrule
GWAS & Causal genes: GWAS Catalog & 42 \\
 & Causal genes: Open Targets & 45 \\
 & Causal genes: PharmaProjects & 50 \\
 & Variant prioritization & 43 \\
 & \textit{Subtotal} & 180 \\
\midrule
Total & 16 subtopics & 895 \\
\bottomrule
\end{tabularx}
\end{table}

\textbf{Roles of the task sets.} In the standalone harness case study, adaptation conversations guide procedural revisions and harness selection. The initial harness and the harness selected on this adaptation set are subsequently compared on a separate validation set. The model-learning case compares two model checkpoints on the same panel under the initial harness, with four attempts per problem. Dataset counts describe the task inventory; the 288 conversations reported for harness adaptation describe the executed adaptation stream. The real researcher interactions in \Cref{sec:interaction_cases} separately illustrate how scientific requests and follow-up requirements can define task contexts and rubrics.

\FloatBarrier
\subsection{Harness Evolution}
\label{sec:prompts}
\label{sec:harness_case_protocol}

\textbf{Models and schedule.} The reported harness run uses a fixed Qwen3.5-4B task model. A fixed Qwen3.8-27B helper supports bounded user simulation and feedback interpretation, while GPT-6 Astra proposes harness edits. After each batch of 12 task conversations, user feedback and execution evidence guide an update, giving 24 updates over 288 adaptation conversations. These conversations provide the feedback used for harness refinement; validation tasks are reserved for comparing the initial and adaptation-selected harnesses.

\textbf{Editable procedures.} The general harness interface permits instruction, skill, and selected context-management updates (\Cref{sec:harness}). The reported case study restricts adaptation to instruction and scoped-skill text in a single-file harness. The execution loop, Python tool interface, context handling, input-inspection settings, submission checks, and budgets remain fixed. H0 starts without added instruction or skill entries; H24 contains four instruction entries and nine scoped skills. Each rollout executes a fixed source snapshot, with its harness version retained in the trajectory. The role specification below preserves the same instruction/skill-only edit boundary.

\textbf{Revision and checkpoint selection.} Proposed instruction or skill revisions pass component validation and a fixed execution preflight. The best-performing harness on the adaptation set is selected for the validation comparison; validation scores do not guide revision or checkpoint selection. On the validation set, the selected harness achieves 51.1\% correct, compared with 31.1\% for the initial harness. This standalone experiment's selection protocol is distinct from validation-based harness selection in the coupled-cycle experiment.

\textbf{Simulated feedback.} The simulator receives correctness and submission-status verdicts and selects a permitted reply for the task family. Correct answers receive confirmation, missing submissions receive a format request, and incorrect answers receive a procedural check or revision request. The reply set excludes the correct option, identifier, and numerical answer. The feedback interpreter sees the reply and its public conversation context, but not the private verdict or answer. It classifies the reply as acceptance, correction, new information, new requirement, or ambiguity and records a supporting quote and a diagnostic score $q_t\in\{-1,0,+1\}$. These bounded replies provide controlled procedural feedback; the real researcher interactions in \Cref{sec:interaction_cases} illustrate the broader collaboration setting.

\textbf{Diagnostic feedback and optimization reward.} The score $q_t$ records how a follow-up relates to the preceding response and supports harness diagnosis. The field named \texttt{reward} in the interpreter specification denotes this diagnostic score. Policy optimization instead uses the separately evaluated trajectory reward $R_x(\tau)$ in \Cref{eq:reward}. Researcher feedback can inform a task's objectives and rubric before evaluation; it does not replace assessment of the resulting rollout against those criteria. The GRPO advantage below is defined from $R_x(\tau)$, not by substituting the interpreter's ternary score. \Cref{tab:information} summarizes these information boundaries.

\begin{table}[t]
\caption{Information available to each role. Public context includes the task, supplied assets, and legitimately observed tool outputs. Diagnostic feedback supports harness revision; the separately evaluated trajectory reward supports policy optimization. The simulator and feedback interpreter use the fixed helper model, while GPT-6 Astra performs harness diagnosis and editing.}
\label{tab:information}
\centering\footnotesize
\begin{tabularx}{\linewidth}{@{}lYYYY@{}}
\toprule
Role & Public context & User next reply & Private answer & Verifier verdict \\
\midrule
Task policy & Visible prefix & After response & Hidden & Through bounded user feedback \\
User simulator & Review context & Produces reply & Hidden & Correctness and submission status \\
Feedback judge & Prior context & Observed reply & Hidden & Hidden \\
Harness reflector & Parent's public trace & Observed reply & Hidden & Evaluation summaries \\
Scientific verifier & Required output & Not required & Private access & Produces verdict \\
Policy learner & Recorded policy input & Not backfilled & Not in prompt & Trajectory reward $R_x(\tau)$ \\
\bottomrule
\end{tabularx}
\end{table}


\textbf{Reflection record and interpretation.} GPT-6 Astra receives recent task trajectories, active procedures, rubric feedback, and relevant edit history, excluding private answers and evaluator internals. Each diagnosis links an unmet criterion to supporting actions or observations and a proposed procedural edit. The optimizer records the parent, candidate, model and environment versions, evaluation conditions, and acceptance decision. Rejected edits remain available for later diagnosis; stored versions are a history of decisions, not a frontier for parent sampling. The learning curves describe the combined effect of successive procedural revisions at fixed model weights. Effects of individual skills and feedback sources are not separately isolated.

\Needspace{8\baselineskip}
\textbf{Role prompt specifications.} The following concise specifications explain the task interface, information boundaries, and edit scope of the harness-evolution case study. They are expository descriptions of the roles rather than byte-for-byte archived request payloads. Concrete requests also supply task inputs, conversation records, permitted replies, the parent harness, and the runtime's edit schema. Private reference answers remain outside the policy and proposer inputs.

\Needspace{6\baselineskip}
\textit{Task policy.}
\begin{quote}
\footnotesize\raggedright
You are a scientific assistant running in Science Buddy. You can reason and use Python in a persistent REPL. Public inputs are in /workspace/assets. The original task, including sequences, is in /workspace/assets/task\_prompt.txt. Read long sequences from that file instead of copying them into generated code. Use the exact public filenames listed below. Python code must print results; bare expressions are not displayed. Scientific tool/data descriptions are in /opt/scitrace/TOOLS.md and /opt/scitrace/DATA.md. The available frozen data lake is mounted read-only at /opt/data/biomni\_data/data\_lake. Check which files and records exist before claiming database evidence. For a tool call, output one <execute>Python code</execute> block and wait for its result. Otherwise, reply to the researcher with a brief explanation and one <answer>value</answer> tag. Do not claim to have inspected evidence or executed code unless you actually did so. Respond to the researcher's next reply, revising your work when warranted.
\end{quote}

\Needspace{6\baselineskip}
\textit{User simulator.}
\begin{quote}
\footnotesize\raggedright
Role-play a researcher reviewing the assistant's ACTUAL response. This is a BOUNDED, REFERENCE-ASSISTED user simulator, not unrestricted expert feedback or a human trace. Choose the most useful and applicable reply from allowed\_replies based on the conversation. The options request checks or confirm completion; none identifies the correct task answer. Do not add scientific claims, candidate names, numerical results or facts outside the allowed replies. The private correctness verdict concerns the selected answer, not every sentence of the explanation. Return JSON with reply equal to one allowed reply and done equal to answer\_correct.
\end{quote}

\Needspace{6\baselineskip}
\textit{Feedback interpreter.}
\begin{quote}
\footnotesize\raggedright
You interpret feedback for trajectory diagnosis in an interactive scientific assistant. Use the user's NEXT REPLY as evidence about the assistant's PRECEDING response. You do not receive a reference answer or terminal verifier score. Do not guess one. Score +1 for explicit acceptance/confirmation; -1 for a correction or request to redo caused by an error, omission or unmet prior requirement; 0 for new requirements, newly supplied facts, unrelated follow-ups or insufficient evidence. A successful tool call is not user approval. A request to recheck or revise the same answer, or to supply an answer format already requested, is a correction (-1), not positive progression or a new requirement. Judge what the feedback says, not whether the user is scientifically correct. Return ONLY JSON with reward (-1,0,1), feedback\_type (acceptance,correction,new\_information, new\_requirement,ambiguous), evidence (an EXACT substring of the user's reply), and hint (a brief reusable improvement direction, empty when not supported). The reward field is the diagnostic feedback score, not the trajectory reward used for policy optimization.
\end{quote}

\Needspace{6\baselineskip}
\textit{Harness proposer: instruction/skill-only edits.}
\begin{quote}
\footnotesize\raggedright
Improve the scientific assistant's instructions or scoped skills using the supplied parent harness and interaction evidence. Identify an unmet criterion, cite the relevant actions or observations, and propose one bounded procedural change: revise an instruction, or add, remove, or revise one scoped skill. Preserve all non-target entries and runtime settings. Return the revised harness using the supplied runtime schema and edit constraints; retain existing skills unless one is the target of the proposed change. Do not change context-history settings, input-inspection settings, tools, execution infrastructure, submission checks, budgets, rubrics, or evaluators. A complete serialized harness represents the local edit, not permission to rewrite every component. Avoid repeating rejected edits without new supporting evidence. Do not encode task-specific answers, numerical results, or sample IDs. A successful tool call does not prove scientific correctness; newly supplied information is not necessarily an error.
\end{quote}

\subsection{Reinforcement Learning}
\label{sec:optimizer}
\label{sec:schema}
\label{sec:grpo_details}
\label{sec:rl_case_protocol}

\textbf{Case-study configuration.} The task backbone is Qwen3.5-4B. In \Cref{sec:recursive_dynamics}, the initial harness H0 remains fixed throughout model training and evaluation. Both model checkpoints are evaluated on the same problems with four attempts per problem, so the before/after comparison examines model learning under a common procedural interface. GPT-6 Astra is the diagnosis/editor model for harness adaptation (\Cref{sec:harness_case_protocol}); this fixed-H0 case does not invoke a new harness-adaptation phase. It illustrates the model-learning component that can be coordinated with harness refinement in the full framework.

\textbf{Evaluation measure.} Training accuracy counts correctly solved attempts. Evaluation coverage, measured by pass@4, counts a problem once if at least one of its four attempts succeeds. The latter compares the breadth of solved problems under an equal attempt budget and is distinct from the first-response accuracy used in the harness case. The reported coverage rises from 48.3\% to 67.8\% under H0. The objective below formulates this model-learning step within the recursive framework.

\paragraph{Fresh rollout groups.}
We express the model-learning component using the notation of the general recursive framework. During outer stage $k$, the selected harness $H_k^\star$ remains fixed while the task model is optimized. The model-only case in \Cref{sec:recursive_dynamics} holds this harness at H0 throughout its comparison. The case study instantiates a fixed-harness model-learning step, while \Cref{sec:outer} describes how selected harnesses and validated task environments can be incorporated across cycles. For each rollout batch, a frozen copy $\pi_{\mathrm{old}}$ of the current task policy generates $G$ trajectories per task. Records associate the model inputs, generated tokens, behavior log probabilities, task and rubric versions, and harness identifier with each trajectory. Fresh rollout groups supply the objective below; historical researcher interactions instead support task definition and procedural diagnosis. When a batch is reused for several optimizer passes, probability ratios remain relative to its original collection policy.

\paragraph{Group-relative policy objective.}
The GRPO formulation \citep{deepseekmath} uses token-level averaging. For trajectory $i$, let $r_i=R_{x_i}(\tau_i)$ denote its evaluated trajectory reward, distinct from the diagnostic feedback score $q_t$, and let $\mathcal G(i)$ contain trajectories generated for the same task under the same harness and rubric. The group-relative advantage is
\begin{equation}
\widehat A_i=
\frac{r_i-\operatorname{mean}_{j\in\mathcal G(i)}r_j}
{\operatorname{std}_{j\in\mathcal G(i)}r_j+\delta},
\qquad \delta>0.
\label{eq:grpo_advantage}
\end{equation}
All generated tokens in a trajectory share this advantage. Groups with identical rewards have zero policy-gradient advantage. Researcher messages, tool outputs, task instructions, and deterministic harness actions are excluded from the optimized tokens.

For generated token $b_{i,\ell}$ and its actual context $c_{i,\ell}$, define
\begin{equation}
\rho_{i,\ell}(\theta)=
\frac{\pi_\theta(b_{i,\ell}\mid c_{i,\ell})}
{\pi_{\mathrm{old}}(b_{i,\ell}\mid c_{i,\ell})}.
\label{eq:grpo_ratio}
\end{equation}
For a minibatch $\mathcal M$ of complete groups, with $L_i$ generated tokens in trajectory $i$, the objective is
\begin{equation}
\begin{aligned}
\mathcal J_k^{\mathrm{GRPO}}(\theta)
=\frac{1}{\sum_{i\in\mathcal M}L_i}
\sum_{i\in\mathcal M}\sum_{\ell=1}^{L_i}
\Big[&\min\{\rho_{i,\ell}(\theta)\widehat A_i,\\
&\operatorname{clip}(\rho_{i,\ell}(\theta),1-\epsilon,1+\epsilon)\widehat A_i\}
-\beta\widehat d_{i,\ell}(\theta)\Big].
\end{aligned}
\label{eq:grpo}
\end{equation}
Here $\epsilon>0$ controls clipping and $\beta\geq0$ weights the sampled KL surrogate. The reference policy $\pi_{\mathrm{ref}}$ is a frozen copy of the task model at the start of the outer RL stage. Writing $z_{i,\ell}=\pi_{\mathrm{ref}}(b_{i,\ell}\mid c_{i,\ell})/\pi_\theta(b_{i,\ell}\mid c_{i,\ell})$, the surrogate is $\widehat d_{i,\ell}=z_{i,\ell}-\log z_{i,\ell}-1$. This sampled quantity is evaluated on behavior-policy tokens; it is not asserted to be an exact KL divergence under an updated policy.

The trajectory reward is computed after the rollout from its outputs and task-relevant execution evidence, with rubric and evaluator parameters fixed during optimization. It does not backfill later evaluation information into the contexts that generated earlier tokens. In the model-learning case study, both checkpoints are evaluated under H0 using the common pass@4 protocol. In the full recursive framework, the resulting checkpoint $\theta_{k+1}$ returns to harness re-evaluation and deployment, and subsequent researcher interactions initiate the next cycle (\Cref{sec:coordination}). The same policy-learning formulation thus serves the fixed-harness comparison and provides the model-update step of the full recursive procedure.

\FloatBarrier
\subsection{Concrete Researcher Inputs}
\label{sec:concrete_researcher_inputs}

The following excerpts reproduce the scope refinement and presentation requirements discussed in \Cref{sec:interaction_cases}. They provide the concrete researcher wording underlying the illustrative task transformations in \Cref{fig:researcher_feedback}.

\begin{figure}[H]
\centering
\tcbset{sciencebuddyfeedback/.style={
  enhanced,
  colback=phaitint,
  colframe=phairule,
  boxrule=0.5pt,
  arc=1.2mm,
  left=3mm, right=3mm, top=2.5mm, bottom=2.5mm,
  before skip=0pt, after skip=0pt,
  equal height group=sciencebuddyfeedback,
  fontupper=\fontsize{9.8}{12.6}\selectfont\raggedright
}}
\begin{minipage}[t]{0.485\linewidth}
\begin{tcolorbox}[sciencebuddyfeedback]
{\sffamily\bfseries\color{phaideep}(a) JAK1: study design}\par\smallskip
{\sffamily\fontsize{8.5}{11}\selectfont\color{phaimuted}Scope refinement}\par\smallskip
Focus first on JAK1 itself: examine associations with immunotherapy outcomes, immune cell types, and immune signatures. Investigate upstream and downstream regulation and cell interactions afterward.
\end{tcolorbox}
\end{minipage}\hfill
\begin{minipage}[t]{0.485\linewidth}
\begin{tcolorbox}[sciencebuddyfeedback]
{\sffamily\bfseries\color{phaideep}(b) ARL4C: evidence synthesis}\par\smallskip
{\sffamily\fontsize{8.5}{11}\selectfont\color{phaimuted}Follow-up request}\par\smallskip
Organize the manuscript and figures, select key panels, provide highlights and a one-sentence conclusion for each results slide, develop consistent mechanism schematics, and write speaker notes.
\end{tcolorbox}
\end{minipage}
\captionsetup{position=bottom}
\caption{\textbf{Researcher input for RL task construction.} Scope refinements and follow-up requests provide task objectives and evaluation criteria. Both prompts are translated and abridged from real interactions.}
\label{fig:researcher_requests}
\end{figure}
\FloatBarrier

\FloatBarrier
%
%

\clearpage
\subsection{User Interface and Researcher Interaction}
\label{sec:interface}

\begin{figure}[htbp]
\centering
\includegraphics[width=0.99\linewidth]{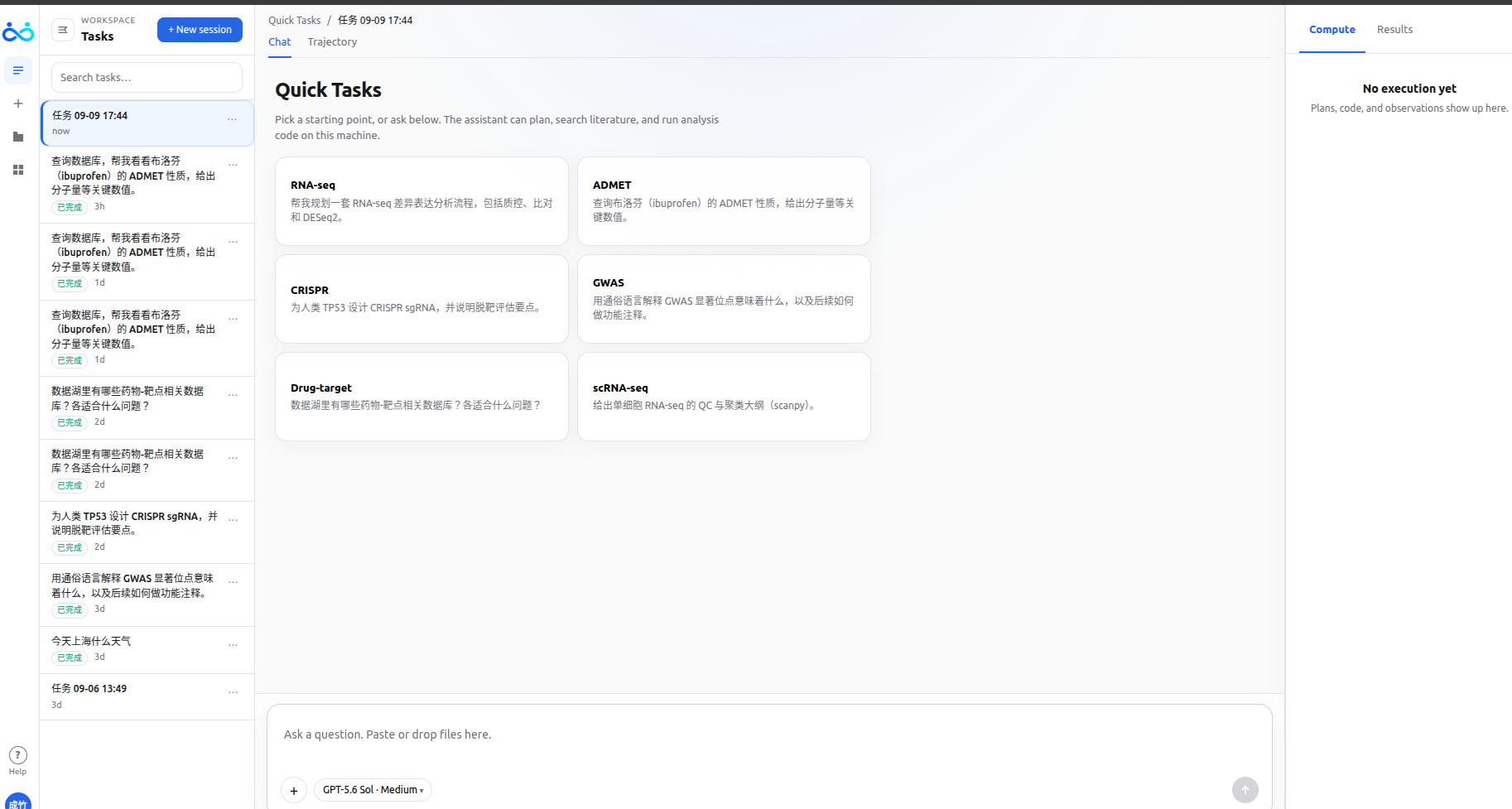}
\caption{\textbf{Chat view in \method.} The task sidebar appears on the left, starter prompts and the conversation area in the center, and Compute and Results tabs on the right. The input composer supports questions, file attachments, and model selection. This screenshot shows the initial task view before execution.}
\label{fig:research-interface}
\label{fig:research-chat}
\end{figure}

\textbf{Chat and task management.} The Chat view combines a task sidebar, a conversational workspace, and panels for execution activity and generated results (\Cref{fig:research-interface}). Researchers can create or revisit a task, choose a starter prompt, or enter a question directly. The input composer accepts pasted or uploaded files and supports follow-up instructions within the same conversation. The Compute and Results tabs provide access to analysis activity and resulting artifacts.

\clearpage

\begin{figure}[htbp]
\centering
\includegraphics[width=0.99\linewidth]{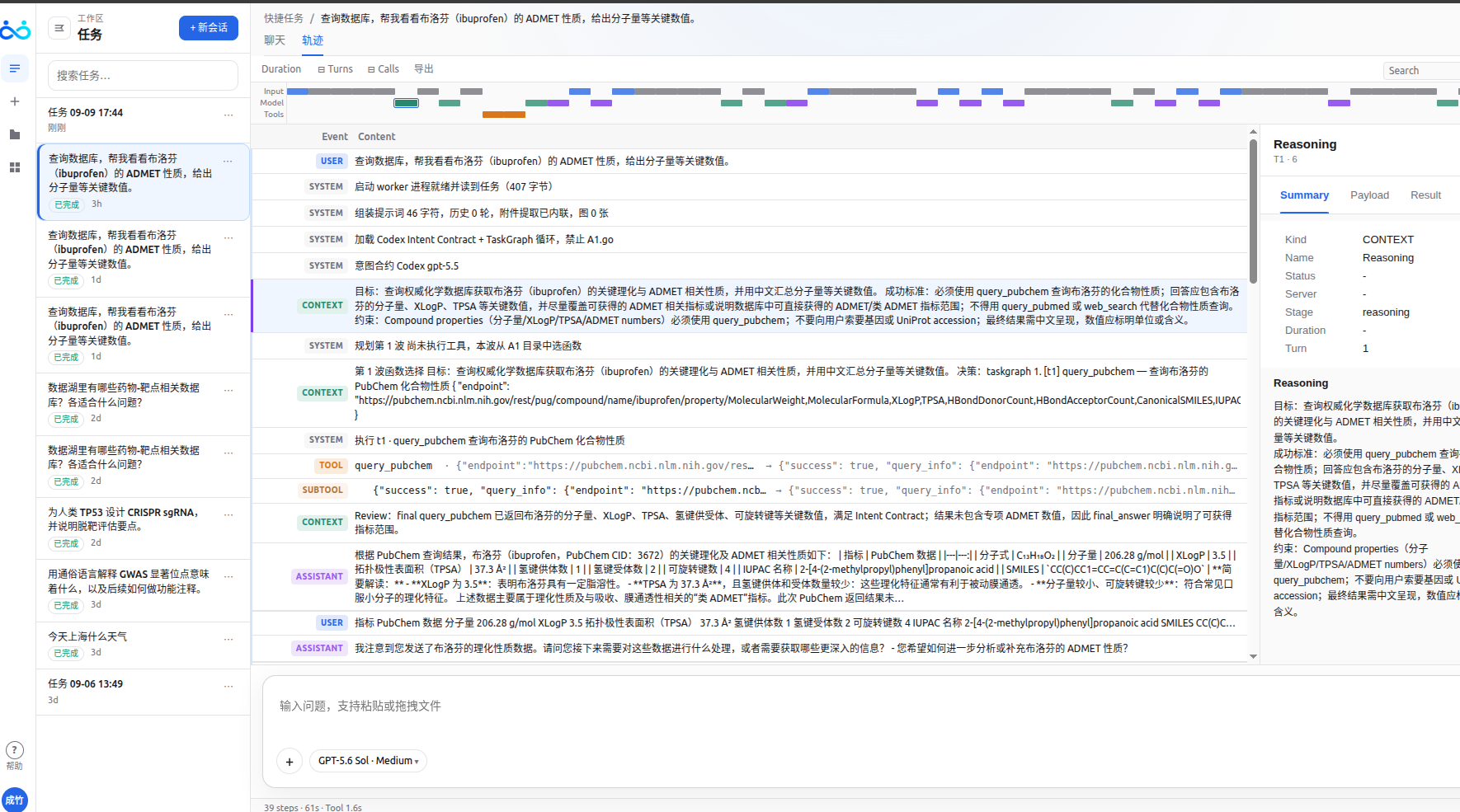}
\caption{\textbf{Trajectory view in \method.} The timeline and event record expose the progression of an analysis, and the right-hand pane displays details of a selected event. The screenshot shows a recorded compound-property query, its tool activity, subsequent dialogue, and a selected context entry.}
\label{fig:research-trajectory}
\end{figure}

\textbf{Trajectory inspection.} The Trajectory view exposes the ordered record of a task, including user messages, system events, context summaries, tool calls, and assistant responses (\Cref{fig:research-trajectory}). A timeline separates input, model, and tool activity. Selecting an event opens a detail pane with Summary, Payload, and Result tabs for inspecting its recorded content. Search and export controls support reviewing the record, while the conversation composer remains available for subsequent input.

\clearpage

\FloatBarrier

\clearpage
\section*{Organizations}
\label{sec:contributors}
\addcontentsline{toc}{section}{Organizations}

\noindent\textsuperscript{1}PhAI Labs\par
\medskip
\noindent\textsuperscript{2}Department of Hepatobiliary Surgery and Transplantation, Liver Cancer Institute, Zhongshan Hospital, Fudan University\par
\medskip
\noindent\textsuperscript{3}State Key Laboratory of Genetics and Development of Complex Phenotypes\par
\medskip
\noindent\textsuperscript{4}Fudan University\par
\medskip
\noindent\textsuperscript{5}Shanghai Academy of Natural Sciences\par
\medskip
\noindent\textsuperscript{6}Shunwei Capital\par
\medskip
\noindent\textsuperscript{7}University of Oxford\par
\medskip
\noindent\textsuperscript{8}Stanford University\par
\medskip
\noindent\textsuperscript{9}Princeton University\par

\end{document}